# Prototypical Learning Guided Context-Aware Segmentation Network for Few-Shot Anomaly Detection

Yuxin Jiang, Yunkang Cao, Weiming Shen, *Fellow, IEEE*

*Abstract*—Few-shot anomaly detection (FSAD) denotes the identification of anomalies within a target category with a limited number of normal samples. Existing FSAD methods largely rely on pre-trained feature representations to detect anomalies, but the inherent domain gap between pre-trained representations and target FSAD scenarios is often overlooked. This study proposes a Prototypical Learning Guided Context-Aware Segmentation Network (PCSNet) to address the domain gap, thereby improving feature descriptiveness in target scenarios and enhancing FSAD performance. In particular, PCSNet comprises a prototypical feature adaption (PFA) sub-network and a context-aware segmentation (CAS) sub-network. PFA extracts prototypical features as guidance to ensure better feature compactness for normal data while distinct separation from anomalies. A pixel-level disparity classification loss is also designed to make subtle anomalies more distinguishable. Then a CAS sub-network is introduced for pixel-level anomaly localization, where pseudo anomalies are exploited to facilitate the training process. Experimental results on MVTec and MPDD demonstrate the superior FSAD performance of PCSNet, with 94.9% and 80.2% image-level AUROC in an 8-shot scenario, respectively. Real-world applications on automotive plastic part inspection further demonstrate that PCSNet can achieve promising results with limited training samples. Code is available at https://github.com/yuxin-jiang/PCSNet.

## I. INTRODUCTION

Anomaly detection aims at recognizing samples that deviate significantly from normal patterns [1], presenting opportunities for automatic industrial defect detection [2, 3]. Given the resource-intensive and time-consuming collection for mass categories [4] in real-world industrial inspection scenarios, few-shot anomaly detection (FSAD) [5-7] has emerged as a viable solution, focusing on detecting anomalies within target categories with limited normal data for training.

Existing FSAD methods are largely inspired by their counterparts, unsupervised anomaly detection methods, which require a substantial number of normal samples for training [8]. These unsupervised anomaly detection methods typically rely on pre-trained representation spaces, in which normal and abnormal samples are distinguishable. Unsupervised methods model the distribution of normal representations and then compare the testing data with the modeled distribution for anomaly detection [9]. Although unsupervised methods have made significant progress, it is important to note that the data distribution in pre-trained networks differs from the data distribution in industrial settings. This mismatch is commonly referred to as a domain gap issue, which may result in normal samples being distributed not compactly within the representation space [10]. Thanks to the large amount of training samples, unsupervised methods can still faithfully model the normal sample distribution. However, this domain gap can have a negative impact on FSAD scenarios. Specifically, there are limited normal training samples for FSAD. If normal samples are distributed not compactly in the representation space, these available normal samples cannot faithfully represent the overall normal sample distribution [11]. Therefore, directly employing pre-trained representations tends to present insufficient anomaly detection performance because of the domain gap.

Numerous FSAD methods have been proposed to mitigate this domain gap. For instance, PatchCore [12] adeptly addresses biases towards specific ImageNet [13] classes by harnessing mid-level feature representations. RegAD [5] suggests aligning different images into a unified coordinate system to enhance structural and content consistency, thereby improving feature compactness. However, solely learning from normal samples can potentially limit the discriminative capabilities of anomaly detection models [14, 15]. As illustrated in Fig. 1(a), when anomalies are not present during training, normal samples cannot be completely distinguished from abnormal samples during testing.

Therefore, we opt to introduce synthetic anomalies to train discriminative FSAD models with the objective of compacting normal features while separating anomalies. To this end, this paper introduces a Prototypical learning guided Context-aware Segmentation Network (PCSNet), comprising a prototype feature adaption (PFA) sub-network and a context-aware segmentation (CAS) sub-network (Fig. 1(b)). The PFA sub-network leverages a pre-trained neural network and a feature adapter to create a compact subspace, thereby preventing anomalies from being distributed similarly to normal samples.

This work was supported in part by the Fundamental Research Funds for the Central Universities of China under Grant HUST:2021GCRC058 *(Corresponding author: Weiming Shen).* The authors are with the State Key Laboratory of Digital Manufacturing Equipment and Technology, Huazhong University of Science and Technology, Wuhan 430074, China (e-mail: yuxinjiang@hust.edu.cn; cyk_hust@hust.edu.cn; wshen@ieee.org)

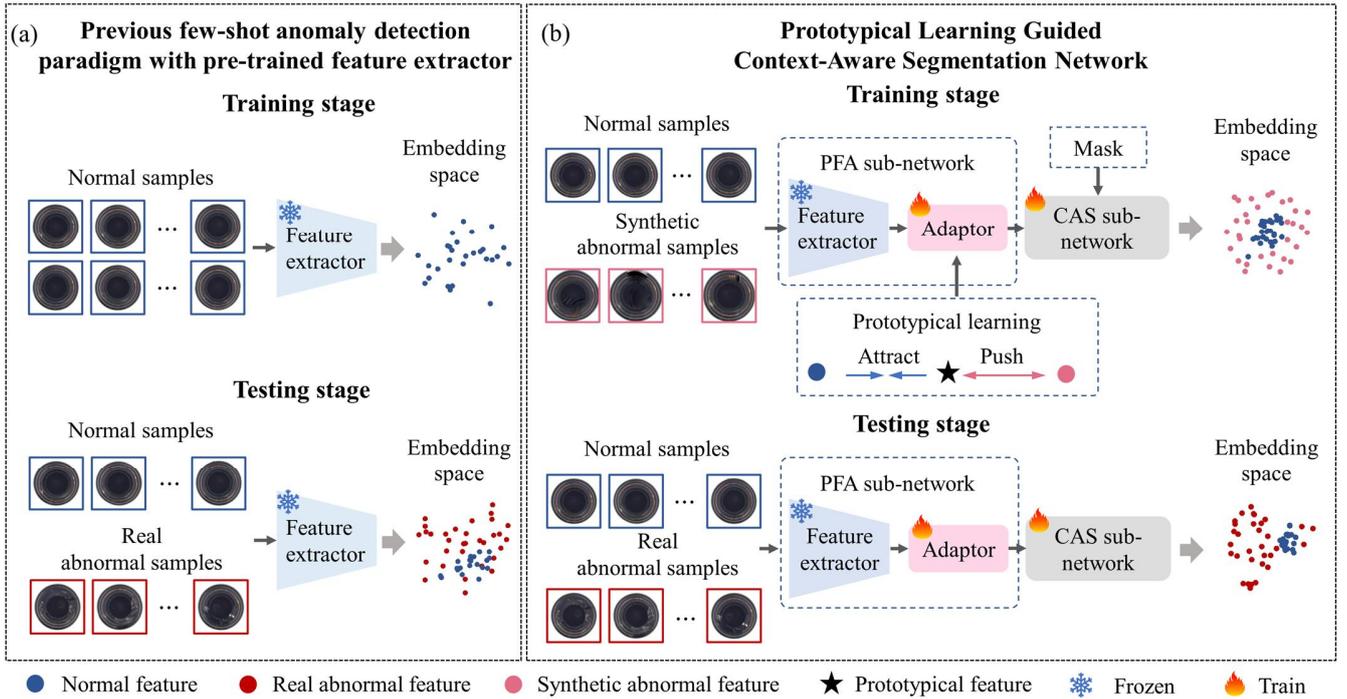

**Fig. 1.** (a): Utilizing only a pre-trained module as the feature extractor often leads to a scattered embedding space for normal samples. As a result, in the testing phase, normal and abnormal features frequently intertwine within the embedding space. (b): PCSNet integrates an adaptor module that employs prototypical learning strategies for adapting the pre-trained features to the target domain. This adaptation process results in the creation of a highly discriminative embedding space. Consequently, in the testing phase, the abnormal and normal features in the embedding space become more distinct.

Specifically, a "normal features compacting (NFC)" loss is developed to establish a decision boundary with the center of prototypical features by promoting compactness within normal features. Then, an "abnormal features separation (AFS)" loss is designed to exploit synthetic anomalies, further compressing the normal feature space by pushing these anomalies beyond a specified margin region from the established decision boundary. A "pixel-level disparity classification (PDC)" loss is additionally utilized to optimize samples that may lead to unreliable results, *i.e.*, hard samples. With the three loss functions, PFA adapts the pre-trained network to the target category, effectively mitigating the domain gap. Then with these adapted features and the matched prototypes, the CAS sub-network is introduced to leverage contextual information to predict pseudo anomaly masks, resulting in a high-precision per-pixel anomaly detection map.

The contributions of this paper are as follows:

- This study presents a PFA sub-network for learning discriminative features. It employs contrastive learning to compress normal features while separating anomalies. The PDC loss amplifies subtle local pixel-wise discrepancies between abnormal and normal features, particularly for hard samples.

- This study constructs a CAS sub-network that uses pseudo masks to effectively extract spatial and semantic characteristics from abnormal features. By exploiting the variations and complexities exhibited by synthetic anomalies, the CAS sub-network enhances the precision of anomaly localization.

- With these two sub-networks, the proposed PCSNet achieves better FSAD results on MVTec and MPDD. Furthermore, in real-world applications such as automotive plastic part inspection, PCSNet has exhibited a noteworthy 92.8% AUPRO, demonstrating that PCSNet can be a viable solution for few-shot scenarios.

The remaining sections of this paper are structured as follows: Section II presents a comprehensive literature review of the field. Section III provides a detailed explanation of the proposed PCSNet method. Extensive evaluation results obtained from rigorous assessments on two publicly available datasets and a real-world application on automotive plastic parts are presented and analyzed in Section IV. Finally, Section V concludes the paper and discusses potential directions for future research.

## II. RELATED WORK

### A. Anomaly Detection

Current anomaly detection methods predominantly concentrate on the unsupervised context due to the limited availability of abnormal samples in practical applications. These unsupervised techniques aim to identify data instances that exhibit significant deviations from expected or normal patterns, without relying on abnormal samples for training [5, 16]. Among these methods, reconstruction-based approaches aim to map input data to a low-dimensional representation and reconstruct the original data from this representation [17]. The underlying assumption here is that training exclusively on normal samples for accurate reconstruction will result in inaccurate reconstruction of abnormal samples during testing [18-21]. This emphasizes the importance of reconstruction errors in anomaly detection. Another approach involves using pre-trained neural networks to model the distribution of normal features [12, 22]. This approach calculates the discrepancies between the distributions of test images and normal features using metrics such as the Mahalanobis distance, thereby identifying anomalies [23]. More recently, knowledge distillation-based methods have gained popularity, which utilize a knowledgeable teacher network as a feature extractor and an unskilled student network trained solely for reconstructing normal data [24-27]. In some cases, multiple student networks were employed to identify anomalies by combining the prediction uncertainties from all students and the regression error of each student compared to the teacher network [28].

*B. Few-shot Learning*

Few-shot learning is a rapidly advancing research domain with the aim of developing models capable of generalizing and predicting new classes using only a few examples from target categories [29]. The prevailing methods predominantly revolve around metric learning and meta-learning. Metric learning [30, 31] focuses on the creation and optimization of distance metrics or similarity measures to accurately differentiate between various samples in a given dataset. Conversely, meta-learning approaches [31, 32] tackle few-shot learning by defining specific optimization or loss functions that facilitate rapid adaptation to new classes. Among these methods, prototypical networks [33-35] have gained significant attention for their ability to extract representative features. Prototypical learning involves extracting and condensing complex object features from support images into highly informative prototypical embedded features. These prototypical features encapsulate the essential characteristics of each class and serve as reference points for classification. During the classification phase, the features extracted from query images are compared with these prototypical features to determine their category [29]. However, some methods focus solely on optimizing relative relations between different classes, *i.e.*, inter-class distances, using cross-entropy loss, which neglects the actual distances between pixels and prototypes, *i.e.*, intra-class compactness [36]. In contrast, our approach employs contrastive learning to not only enhance the separation between normal and abnormal samples, thereby increasing inter-class distances but also to encourage the close proximity of normal samples to each other, promoting intra-class compactness.

*C. Few-shot Anomaly Detection*

Few-shot anomaly detection seeks to address the challenge of limited training samples [7]. In many practical scenarios, traditional anomaly detection methods struggle to accurately differentiate between normal and abnormal data due to the scarcity of training data. In contrast, few-shot anomaly detection overcomes this limitation by enabling effective training of models with smaller datasets. The essence of few-shot anomaly detection lies in leveraging a small number of normal samples to learn representative features of normality. A notable method, TDG [6], leverages image transformations and scale-specific discriminators to enhance the model's representation capability. Another approach, RegAD [5], employs registration-based proxy tasks for representation learning, aligning samples of the same semantic category through affine transformations of feature mappings. GraphCore utilizes a vision isometric invariant graph neural network to extract rotation-invariant structural features, thereby enhancing its ability to distinguish anomalies [37]. However, the absence of abnormal samples during training limits the compactness of normal distributions. To address this challenge, this study introduces PCSNet, a novel framework that combines a PFA sub-network and a CAS sub-network for anomaly detection. The PFA sub-network effectively exploits inherent characteristics in both normal and synthetic abnormal images, resulting in a feature space with strong discriminative properties. Additionally, the CAS sub-network incorporates contextual information to further enhance the accuracy of anomaly detection.

III. PROPOSED METHOD

*A. Problem Definition and Method Overview*

**Problem Statement.** The primary objective of this study is to develop a model capable of effectively detecting anomalies, even when trained on a small training set, denoted as $D_{train}$ with a limited number of normal samples ( $\leq 8$ in this study). After training, the model should demonstrate efficient adaptation capabilities to accurately score the degrees of anomalies and precisely localize the abnormal regions in unseen images from the testing dataset, referred to as $D_{test}$.

**Overview of the proposed Method.** The overview of the proposed PCSNet is depicted in Fig. 2. This study constructs a prototypical feature adaptation (PFA) sub-network that adapts pre-trained features to the target domain, intending to learn compact and discriminative normal feature space for determining anomalies. Subsequently, a context-aware segmentation (CAS) sub-network utilizes the joint contextual information from extracted features and their similarity maps with prototypical features to perform anomaly localization.

*B. Prototypical Feature Adaption Sub-network*

**PFA sub-network design.** The PFA sub-network comprises a pre-trained feature extractor $E$ and a feature adaptor $A$. The input to the PFA sub-network consists of a set of normal images, denoted as $x^n$, randomly selected from $D_{train}$. For each normal sample, a corresponding synthetic abnormal sample, represented as $x^a$, is generated using the NSA method [42]. Specifically, patches of various sizes are cropped from an image in $D_{train}$ and seamlessly integrated into $x^n$ through Poisson image editing, resulting in the synthetic abnormal image $x^a$. The PFA sub-network exploits the pre-trained feature extractor $E$ to extract descriptive features for these input images. However, due to the domain gap between pre-trained datasets, *e.g.*, ImageNet, and target images, these features exhibit insufficient descriptiveness. To address this issue, this study further introduces the feature adaptor $A$ to adapt these pre-trained features to the target domain using the proposed three loss functions. $A$ is composed of three 1x1 convolution layers and a single CoordConv layer [43]. The convolutional layers are employed to extract semantic features, while the CoordConv layer enhances spatial awareness.

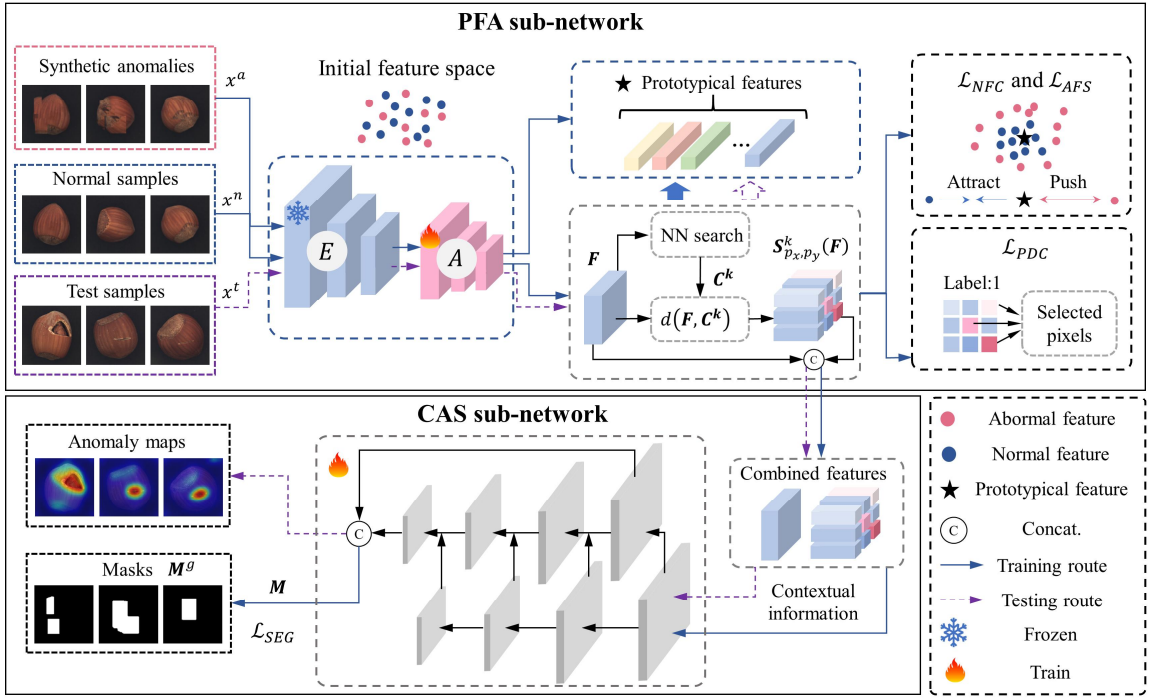

**Fig. 2.** The diagram of the proposed PCSNet model. Firstly, the PFA sub-network extracts discriminative features with three loss functions. Subsequently, the CAS sub-network performs anomaly localization by concatenating the extracted features and similarity maps.

**Normal features compacting loss** is based on the principle of soft-boundary regression, aiming to identify a decision boundary with the smallest radius to densely compact normal features. This loss function prevents anomalies from being closely mapped to normal feature space, thereby aiding in the detection of anomalies that deviate from the normal distribution. Initially, the normal images $x_n$ are processed through the PFA sub-network, resulting in feature patches $F_j^n = A(E(x))$, where $j \in \{1,2,\cdots H \times W\}$. Here, $H$ and $W$ represent the height and width of the features extracted by $E$, respectively. The average of a batch of feature patches serves as the prototype set $\mathbb{C}$. Then, a nearest neighbor search is performed on the feature patches and $\mathbb{C}$ to find the top $k$ nearest prototypical features, represented as $C_j^k \in \mathbb{C}$. Finally, $F_j^n$ is embedded close to $C_j^k$ within a predefined radius $r$. This radius $r$ functions as a threshold for determining the abnormality of feature patches.

$$\mathcal{L}_{NFC} = \frac{1}{J \times K} \sum_{j=1}^{J} \sum_{k=1}^{K} max\{0, \mathcal{D}(F_j^n, C_j^k) - r^2\} \quad (1)$$

where $\mathcal{D}(.,.)$ is a predefined distance metric, *i.e.*, Euclidean distance in this study, to quantify the differences and similarities among samples in the feature space. The number of feature patches, denoted as $J$, is equal to $H \times W$.

**Abnormal features separation loss.** The presence of significant inter-class variations among normal samples can lead to unoccupied regions between the normal and prototypical features. Consequently, the loose normal feature space may accommodate certain abnormal samples that exhibit similarities to normal instances. To increase the compactness of normal space, the intuitive idea is to use synthetic anomaly that compresses the space by maintaining distance from the prototypical features. Specifically, the loss function is defined as follows:

$$\mathcal{L}_{AFS} = \frac{1}{J \times K} \sum_{j=1}^{J} \sum_{k=1}^{K} max\{0, (r+\alpha)^2 - \mathcal{D}(F_j^a, C_j^k)\} \quad (2)$$

where $\alpha$ is a radius relaxation coefficient, which prevents compelling anomalies from deviating extensively from the normal distribution. Instead, it solely pushes the anomalies outside the margin region, thereby mitigating the risk of overfitting the model to synthetic anomalies.

**Pixel-level Disparity Classification Loss** is an additional function designed to effectively differentiate abnormal features from normal ones, especially for "hard samples". In some cases, abnormal samples may bear a striking resemblance to normal samples, which poses a significant challenge in distinguishing them. Such anomalies are referred to as "hard samples". To address this challenge, this study proposes a pixel-level disparity classification loss function (PDC) that focuses on pixels exhibiting the greatest disparities when compared with normal samples. To achieve this, similarity maps are calculated by comparing the extracted features $F^n$ and $F^a$ with prototypical features. These maps quantify the similarity between each pixel and a normal reference, which is formulated as follows:

$$S_{p_x,p_y}^k(F) = \mathcal{D}(F_j, C_j^k) \quad (3)$$

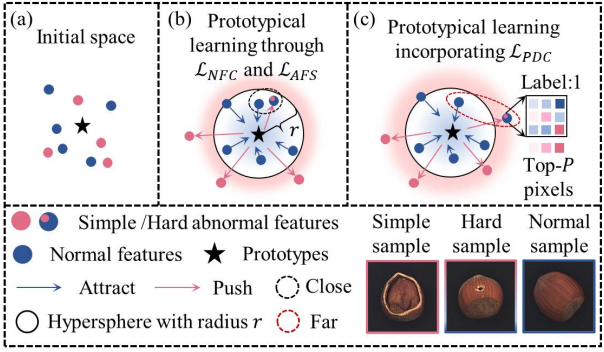

**Fig. 3.** Visual representation of prototypical learning process. (a) In the initial stage, both abnormal and normal data display a dispersed distribution in the embedding space. (b) By employing the NFC and AFS losses, there is a gradual segregation of abnormal and normal data, although certain anomalies (hard samples) with a similar visual appearance to normal samples still exhibit proximity to the normal samples. (c) The PDC loss helps to identify specific local pixels that exhibit the most significant differences when compared with normal samples. This allows for the separation of hard samples from normal ones by concentrating on these disparities.

where, $p_x, p_y$ indicate the specific location of a pixel $p$, and $F \in \{F^n, F^a\}$.

We then identify the top-$P$ pixels with the greatest disparity values in $S_{p_x,p_y}^k(F)$. This selected set of $P$ vectors $\Psi(F)$ is defined as:

$$\Psi(F) = \arg\max\left(S_{p_x,p_y}^k(F), P\right) \quad (4)$$

where, $\arg\max(\cdot,\cdot)$ denotes the selection of the top-$P$ pixels with the largest disparity values.

Subsequently, the average value of these selected pixels within $\Psi(F)$ is utilized as the foundation for image classification, employing the PDC loss function denoted as $\mathcal{L}_{PDC}$:

$$\mathcal{L}_{PDC} = \mathcal{L}(\mathcal{G}_s(F), y_x) \quad (5)$$

where $y_x$ represents the assigned label for $x$. Specifically, $y_x = 1$ if $x$ is an anomaly; $y_x = 0$ if $x$ is a normal sample. Furthermore, $\mathcal{G}_s(F)$ is defined as:

$$\mathcal{G}_s(F) = \frac{1}{P}\sum_{p\in\Psi(F_n)}\left(S_{\Psi(F)}^k(F)\right) \quad (6)$$

where, $\mathcal{G}_s(F)$ represents the degree of abnormality of an embedded feature.

By prioritizing specific local pixels over the overall features, the proposed method emphasizes the hard sample with subtle discrepancies with normal features. Fig. 3 provides a visual representation of the entire prototypical learning process, providing deep insights into the unique functions of the three mentioned loss functions.

*C. Context-aware Segmentation Sub-network*

**CAS sub-network design.** This study employs an FPN-like [44] CAS sub-network, denoted as $S$, to generate fine-grained anomaly maps. To effectively detect abnormal regions that vary in magnitude, this study adopts the multi-scale aggregation framework outlined in [29] to extract multi-scale features from $F$, thereby capturing contextual information across different resolutions. These multi-scale features are subsequently upsampled and concatenated with a similarity map $S_{p_x,p_y}^k(F)$ to serve as input, compelling $S$ to learn a distance function between the extracted features and their corresponding prototypes, thereby mitigating the risk of overfitting to synthetic appearances. Following this, convolutional layers are employed to refine the concatenated features, and residual connections are integrated during the pyramid feature processing to enhance feature interaction. Ultimately, the concatenation of features across all scales yields a segmentation result, which is utilized for loss computation.

**Loss functions.** The CAS sub-network employs a segmentation loss function, denoted as $\mathcal{L}_{SEG}$, to facilitate the generation of a pixel-wise anomaly score map, which indicates the likelihood of individual pixels being categorized as anomalies. A simple cross-entropy loss $\mathcal{L}_{SEG}$ is utilized for training the CAS sub-network:

$$\mathcal{L}_{SEG} = \frac{1}{N}\sum_{j=1}^{N}\left(M_j - M_j^g\right)^2 \quad (7)$$

where, $M_j$ and $M_j^g$ are the output anomaly localization and ground truth masks, respectively. Here, $j$ denotes individual pixels out of the entire set of pixels.

The total loss function employed for training PCSNet is expressed as follows:

$$\mathcal{L}_{total} = \mathcal{L}_{NFC} + \mathcal{L}_{AFS} + \lambda_1 \mathcal{L}_{PDC} + \lambda_2 \mathcal{L}_{SEG} \quad (8)$$

where, $\lambda_1$ and $\lambda_2$ are hyper-parameters balancing individual loss functions.

*D. Inference*

During the training phase, the PFA and CAS sub-networks are trained jointly using the total loss function. During the testing phase, the output of the CAS sub-network can be easily interpreted as pixel-wise anomaly scores. This study also applies Gaussian smoothing to the pixel-wise anomaly scores with a standard deviation ($\sigma$) of 4 for post-processing.

IV. EXPERIMENTS AND ANALYSES

*A. Datasets and Evaluation Metrics*

**Datasets.** To evaluate the performance of the proposed PCSNet, this study conducted experiments on two publicly available datasets: MVTec AD [45] and Metal Part Defect Detection (MPDD) [46].

**MVTec AD** is a widely recognized and utilized dataset for anomaly detection. It comprises a total of 5354 high-resolution images, including five texture classes and ten object classes. Each sub-dataset is divided into a training set with only normal samples and a test set with both normal samples and various abnormal samples labeled with pixel-precise annotations.

**MPDD** is a recently introduced dataset specifically designed for defect detection in painted metal part fabrication. It encompasses a diverse range of metal parts, comprising six distinct classes. The dataset captures images under various conditions, including different spatial orientations, positions, distances of multiple objects, varying light intensities, and non-homogeneous backgrounds.

TABLE I Comparison of k-shot anomaly detection and localization results on the MVTec and MPDD datasets: AUC (%). The average and standard deviation are reported for each measurement over five random seeds. The best and second-best results are respectively marked in bold and underlined.

| Data | Method | Backbone | MVTec | | MPDD | |
|---|---|---|---|---|---|---|
| | | | Image-level | Pixel-level | Image-level | Pixel-level |
| 2-shot | RD4AD [38] | WRN50 | 75.5 | 71.8 | 61.8 | 74.5 |
| | CFA [39] | WRN50 | 81.1 | 91.0 | 58.8 | 78.2 |
| | PatchCore [12] | WRN50 | 87.8 | 94.8 | 59.6 | 79.2 |
| | RegAD [5] | Res18 | 85.7 | 94.6 | 63.4 | 93.2 |
| | RFR [40] | Res18 | 86.6 | **95.9** | - | - |
| | PACKD [41] | WRN50 | <u>90.2</u> | 95.0 | <u>66.6</u> | **94.4** |
| | **PCSNet** | WRN50 | **90.4±0.6** | <u>95.7±0.3</u> | **67.7±1.0** | <u>92.7±0.4</u> |
| 4-shot | RD4AD | WRN50 | 76.9 | 72.2 | 62.1 | 75.5 |
| | CFA | WRN50 | 85.0 | 91.3 | 59.3 | 78.7 |
| | PatchCore | WRN50 | 89.5 | 95.0 | 59.8 | 79.8 |
| | RegAD | Res18 | 88.2 | 95.8 | 68.8 | 93.9 |
| | RFR | Res18 | 89.3 | <u>96.4</u> | - | - |
| | PACKD | WRN50 | <u>91.6</u> | 96.2 | <u>69.8</u> | <u>94.8</u> |
| | **PCSNet** | WRN50 | **92.1±0.9** | **96.5±0.2** | **70.8±1.0** | **95.1±0.2** |
| 8-shot | RD4AD | WRN50 | 78.5 | 73.0 | 62.4 | 75.7 |
| | CFA | WRN50 | 90.9 | 91.6 | 60.9 | 79.0 |
| | PatchCore | WRN50 | 94.3 | 95.6 | 60.0 | 80.3 |
| | RegAD | Res18 | 91.2 | 96.7 | <u>71.9</u> | 95.1 |
| | RFR | Res18 | 91.9 | 96.9 | - | - |
| | PACKD | WRN50 | **95.3** | **97.3** | 70.5 | <u>95.3</u> |
| | **PCSNet** | WRN50 | <u>94.9±0.2</u> | <u>97.1±0.2</u> | **80.2±1.2** | **96.2±0.1** |

**Evaluation Metrics.** To assess the performance of our method, this study utilized two widely used metrics in anomaly detection: Image-level Area Under the Receiver Operating Characteristics (AUROC) and Pixel-level AUROC to evaluate anomaly detection and anomaly localization performance, respectively.

**Implementation Details.** This study utilizes WideResNet50 [47] as the feature extractor network by default. Adam optimizer is utilized with a learning rate of 0.001 for the feature adaptor in the PFA sub-network and 0.0001 for the CAS sub-network. Both the networks are trained for 50 epochs. The hyperparameter $K$ in Eq. (1) is fixed at 3 and parameter $P$ in Eq. (4) is set to 50. $\lambda_1$ and $\lambda_2$ are set to 30 and 150 respectively. The implementation was carried out using PyTorch V1.12.0 framework, and the models were trained on a Nvidia GeForce RTX 3090 Ti GPU and an Intel i9@3.00GHz CPU.

*B. Anomaly Detection and Localization*

Table I shows FSAD averaged performance on both MVTec and MPDD datasets. The proposed PCSNet demonstrates competitive results to other FSAD alternatives. Specifically, for anomaly detection, PCSNet exhibits improvements of 4.7%, 3.9%, and 3.7% in RegAD, and 9.3%, 7.1%, and 4.0% in CFA across 2-, 4-, and 8-shot scenarios on the MVTec dataset. Additionally, PCSNet exhibits remarkable performance on the MPDD dataset, achieving an accuracy of 80.2% in the 8-shot scenario, which surpasses PACKD by 9.7% and outperforms RegAD by 8.3%. Furthermore, regarding anomaly localization, PCSNet yields competitive results, ranking first and second for both datasets. To further assess the effectiveness of PCSNet in anomaly localization, this study conducted a visual analysis of several cases from the MVTec and MPDD datasets. As shown in Fig. 4, our method showcases the best anomaly localization performance. PCSNet accurately localizes entire abnormal regions in images containing significant anomalies, such as hazelnuts and cables. Furthermore, our method excels in precisely localizing relatively smaller anomalies, such as samples of capsules, pills, and screws. Additionally, even in images featuring multiple abnormal regions, such as the grid and toothbrush, our method successfully localizes all abnormal regions without any omissions. While other FSAD alternatives tend to localize anomalies within a coarse range, leading to the misclassification of many normal regions, our proposed method notably enhances the precise localization of anomalies.

*D. Ablation Study*

To quantitatively evaluate the influence of various components in our methodology, we conducted a comprehensive set of experiments on the MVTec and MPDD datasets. The ablation results of our method under different $k$-shot settings are shown in Table II, where "$S$" represents the CAS sub-network, "$\mathcal{L}_{NFC} + \mathcal{L}_{AFS}$" refer to NFC loss and AFS loss, and "$\mathcal{L}_{PDC}$" represents the PDC loss function. Here are the key findings:

**Loss functions Analysis.** As shown in Table II, 1) Without all three loss functions (fourth row), the method only achieves 71.9%/68.8%/72.1% on the MVTec dataset and 50.4%/55.0%/55.5% on the MPDD dataset. 2) Adding only the PDC loss function (fifth row) promotes the performance dramatically by 12.0%/15.8%/13.4% on the MVTec dataset, and 3.1%/0.5%/2.4% on the the MPDD dataset. 2) Additionally, with the loss functions "$\mathcal{L}_{NFC} + \mathcal{L}_{AFS}$" (sixth row), the performance on the MVTec and MPDD datasets is further

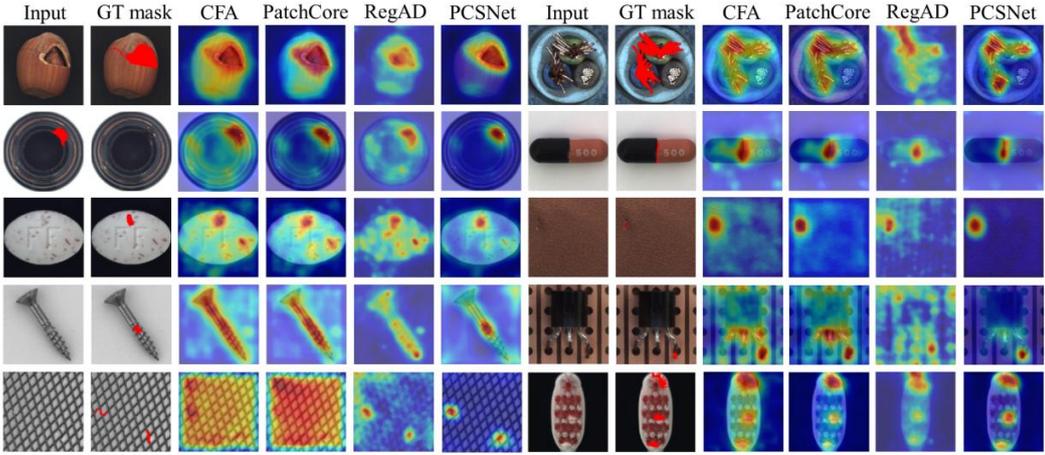

**Fig. 4.** The qualitative results of anomaly localization for PCSNet on various categories from the MVTec datasets. The evaluated categories include hazelnut, bottle, pill, screw and grid.

improved by 6.5%/7.5%/9.4%, and 14.2%/15.3%/22.3%, respectively.

TABLE II ABLATION STUDIES OF DIFFERENT MODULES ON MVTEC AND MPDD FOR ANOMALY DETECTION WITH $k = 2, 4, 8$. THE BEST RESULT IS HIGHLIGHTED IN BOLD.

| Modules | | | | MVTec | | | MPDD | | |
|---|---|---|---|---|---|---|---|---|---|
| $\mathcal{L}_{NFC} + \mathcal{L}_{AFS}$ | $\mathcal{L}_{PDC}$ | $S$ | | 2-shot | 4-shot | 8-shot | 2-shot | 4-shot | 8-shot |
| ✓ | | | | 82.6 | 86.0 | 85.4 | 53.9 | 70.9 | 72.5 |
| | ✓ | | | 79.9 | 84.1 | 86.8 | 49.3 | 56.6 | 54.2 |
| ✓ | ✓ | | | 86.5 | 89.2 | 91.0 | 64.3 | 75.1 | 77.1 |
| | | ✓ | | 71.9 | 68.8 | 72.1 | 50.4 | 55.0 | 55.5 |
| ✓ | | ✓ | | 83.9 | 84.6 | 85.5 | 53.5 | 55.5 | 57.9 |
| ✓ | ✓ | ✓ | | **90.4** | **92.1** | **94.9** | **67.7** | **70.8** | **80.2** |

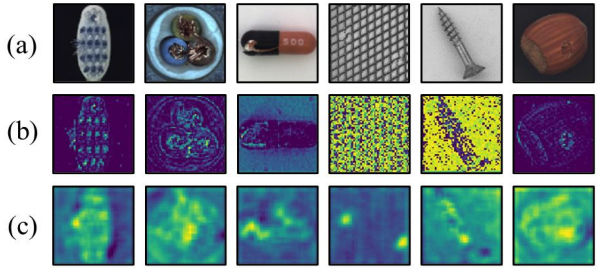

**Fig. 5.** The visualizations of extracted features. (a) represents input images. (b) displays the features obtained solely from $E$, while (c) represents the extracted features acquired by the PFA sub-network, which was trained using both "$\mathcal{L}_{NFC} + \mathcal{L}_{AFS}$" and "$\mathcal{L}_{PDC}$" loss functions.

Additionally, Fig. 5 provides insights into the intrinsic characteristics of the extracted features. The features extracted by the frozen network $E$ (Fig. 5 (b)) demonstrate a balanced allocation of attention towards both abnormal and normal features. On the other hand, the features extracted by the PFA sub-network, trained using the combination of loss functions "$\mathcal{L}_{NFC} + \mathcal{L}_{AFS}$" and "$\mathcal{L}_{PDC}$" (Fig. 5 (c)), prominently emphasizes the abnormal features. These observations suggest that the proposed loss functions effectively transfer the pre-trained features from the source domain into the target domain, which facilitates the acquisition of features with enhanced discernibility.

**CAS Sub-network Analysis.** The integration of the CAS sub-network also brings significant improvements in the anomaly detection performance, e.g., improvements of 3.1%/2.4%/4.2% were observed on the MVTec dataset. Fig. 6 presents anomaly scores for each patch feature of capsule-class and toothbrush-class images. The analysis reveals a significant issue arising from the absence of the CAS sub-network: numerous normal points exhibit high anomaly scores, which results in incorrect anomaly classification (refer to Fig. 6 (a) and (b)). However, the introduction of the CAS sub-network substantially mitigates this issue, as demonstrated in Fig. 6 (c) and (d). Furthermore, Fig. 7 provides insights into the comparison of the similarity map and the final score map. The similarity map reveals a significant number of false positive anomaly detections, whereas the score map generated by the CAS sub-network demonstrates a more accurate localization result, which showcases the effectiveness of the CAS sub-network in mitigating misclassifications.

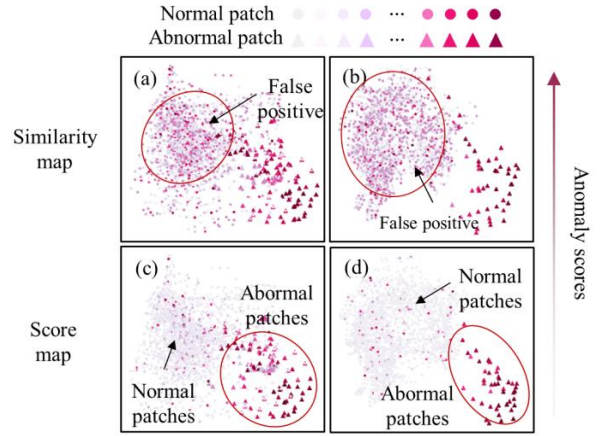

**Fig. 6.** Visualization of similarity maps and score maps of each feature patch of one image from the capsule class (a) and (c), as well as toothbrush class (b) and (d). In the visualization, circles represent the anomaly scores corresponding to normal features, while triangles denote the anomaly scores associated with abnormal features. The intensity of the color used to depict each data point reflects the magnitude of its anomaly score, with darker shades indicating higher anomaly scores.

Additionally, this study explored the impact of different CAS network inputs on anomaly localization. In addition to utilizing a combination of similarity maps ("$S$") and extracted features ("$F$"), the study also investigated the effects of using only "$F$" as input. The experimental results reveal that the combination of "$S$" + "$F$" produced the most optimal outcomes (see Fig. 8), which substantiates the efficacy of employing similarity maps to guide the generation of pixel-level score maps.

Furthermore, to explore the influence of parameter $\lambda_2$ on the CAS Sub-network, this study conducts several experiments using different $\lambda_2$ values while maintaining $\lambda_1$ at 30. As shown in Fig. 9, PCSNet remains nearly stable with an increasing $\lambda_2$ and demonstrates relatively superior results when $\lambda_2$ =150. Therefore, $\lambda_2$ is set to 150 throughout the experimental process.

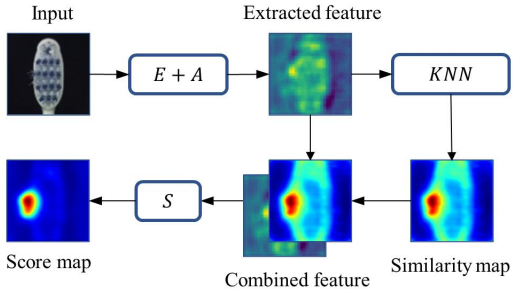

**Fig. 7.** The generation process of a fine-grained score map.

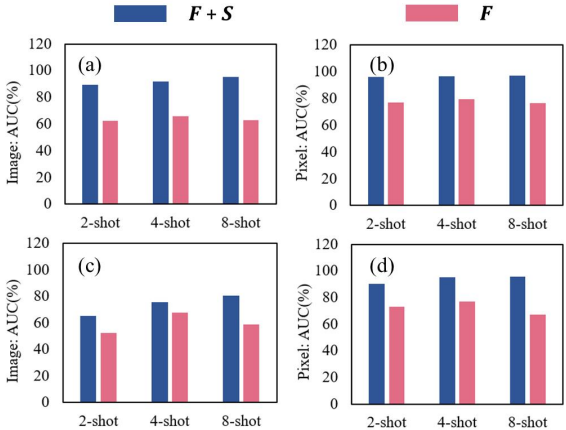

**Fig. 8.** Impact of different inputs of the CAS sub-network on anomaly detection and localization. (a) and (d) displays the anomaly detection and localization results on the MVTec dataset, respectively. (c) and (d) represent the anomaly detection and localization results on the MPDD dataset, respectively. The *x*-axis of both figures represents different shots, specifically 2, 4, and 8, while the *y*-axis represents the corresponding detection or localization results. Abbreviations were used in the legend for convenience. "$S$" denotes the similarity map, "$F$" represents the extracted features.

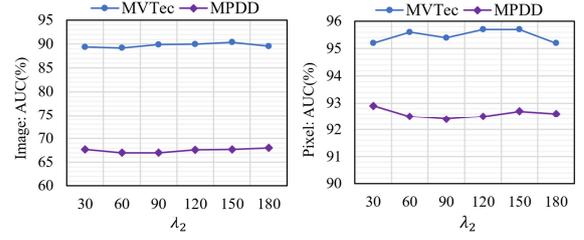

**Fig. 9.** Impact of $\lambda_2$ on anomaly detection and localization.

### E. Feature Distribution Analysis

To assess the efficacy of the proposed PFA sub-network in enhancing anomaly detection performance, we utilize t-SNE for visualizing the learned features on the MVTec dataset, as depicted in Fig. 10. In this figure, blue and red dots represent normal and real abnormal samples from the test set, respectively, while pink dots denote synthetic abnormal samples. The figures show that some of the synthetic anomalies are distributed in the same region as the real anomalies (Fig. 10 (a)), while others are separated into different regions (Fig. 10 (b) and (c)). Nevertheless, the model effectively discriminates normal samples from real anomalies in both scenarios. This suggests that (1) synthetic anomalies and real anomalies share certain similarities while also manifesting differences; (2) Although there are some differences between synthetic anomalies and real anomalies, learning from synthetic anomalies can guarantee a clear separation between normal and abnormal features.

Furthermore, this study also visualizes the original pre-trained features and adapted features of each category on the MVTec dataset. As illustrated in Fig. 11, the adapted features demonstrate a more compact distribution within each category, enabling better separation between features of different categories compared to original pre-trained features This enhancement aids in the detection of real anomalies by focusing on the distinctive features of normal data.

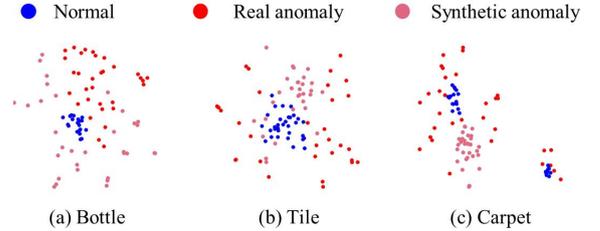

**Fig.10.** t-SNE visualization of features generated by the PFA sub-network.

### F. Complexity Analysis

Table III presents the complexity comparison between several state-of-the-art methods in terms of inference speed and model size. Notably, the inference speed of PCSNet surpasses that of CFA and Patchcore by 1.9× and 3.2×, respectively. Furthermore, certain feature restoration models like SPADE require the storage of a feature gallery, leading to significant memory consumption. In contrast, our approach requires less memory, being 1.8× smaller than SPADE. Although our method entails higher memory consumption than some counterparts, it delivers superior anomaly detection outcomes. In conclusion, PCSNet offers a good trade-off between inference speed, model size, and anomaly detection performance.

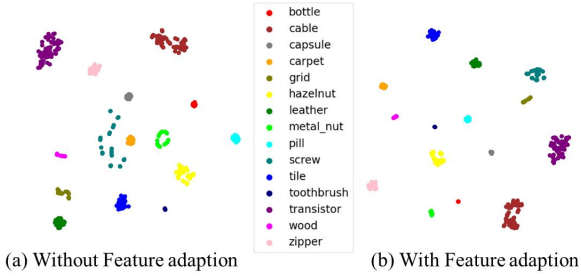

(a) Without Feature adaption  (b) With Feature adaption

**Fig. 11.** t-SNE visualization of features learned from the MVTec dataset, using (a) the baseline without the feature adaption, and (b) the proposed method with the feature adaption.

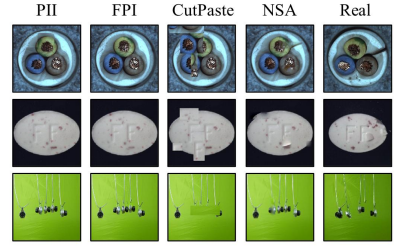

**Fig. 12.** Synthetic anomalies created through the PII, FPI, CutPaste, NSA methods, as well as real anomalies.

TABLE III COMPLEXITY COMPARISON IN TERMS OF INFERENCE SPEED (FPS) AND MODEL SIZE (MB) AND ANOMALY DETECTION PERFORMANCE ON 2-SHOT SCENARIOS (AUC%) ON MVTEC.

| Method | Inference speed | Model size | Performance |
|---|---|---|---|
| SPADE | 0.39 | 1400 | 70.7 |
| RD4AD | 2.95 | 352 | 75.6 |
| CFA | 4.87 | 301 | 81.1 |
| Patchcore | 2.87 | **275** | 87.8 |
| PCSNet | **9.29** | 781 | **90.4** |

*G. The Influence of Introducing Abnormal Samples*

**Synthetic Abnormal Samples.** This experiment aims to investigate the impact of various anomaly synthesis methods on anomaly detection performance. The findings from Table IV show that PCSNet achieves higher accuracy in detecting anomalies when the NSA method is applied to generate synthetic anomalies, as opposed to using the Cutpaste [48], FPI [49], and PII [50] anomaly generation methods. This indicates a positive correlation between the quality of anomaly images (refer to Fig. 12) and the effectiveness of anomaly detection.

TABLE IV ANOMALY DETECTION PERFORMANCE WITH VARIOUS ANOMALY GENERATION METHODS. THE BEST RESULT IS HIGHLIGHTED IN BOLD.

| Dataset | Data | PII [49] | FPI [48] | Cutpaste [47] | NSA [42] |
|---|---|---|---|---|---|
| MVTec | 2 | 85.8 | 88.0 | 87.4 | **90.4** |
|  | 4 | 88.0 | 87.1 | 91.8 | **92.1** |
|  | 8 | 90.8 | 91.6 | 93.4 | **94.9** |
| MPDD | 2 | 51.8 | 57.7 | 67.0 | **67.7** |
|  | 4 | 60.7 | 64.5 | 69.6 | **70.8** |
|  | 8 | 73.6 | 70.9 | 78.7 | **80.2** |

**Real Abnormal Samples.** To further explore the impact of anomaly data quality on anomaly detection performance, this study investigates how the training of PCSNet with a restricted set of real abnormal samples influences its overall effectiveness.

The training dataset encompasses a constrained number of real abnormal samples ($n = 1, 2, 4$) along with a limited number of normal samples. As depicted in Table V, the performance of PCSNet is positively influenced by an increase in the number of real abnormal samples used for training. Specifically, the introduction of a single abnormal sample leads to significant improvements in detection performance compared with the absence of any abnormal samples. For instance, the cable class exhibits a performance increase of 2.2%, the pill class shows a 4.7% improvement, and the connector class achieves a 4.0% enhancement compared with the absence of abnormal samples. These considerable improvements can be attributed to the ability of the network to effectively extract intrinsic characteristics inherent in real anomalies.

TABLE V ANOMALY DETECTION PERFORMANCE WITH LIMITED REAL ABNORMAL SAMPLES ($n = 1, 2, 4$) AND A SET OF 8 NORMAL SAMPLES. THE BEST RESULT IS HIGHLIGHTED IN BOLD.

| Classes | Anomaly type | Real abnormal samples | | | |
|---|---|---|---|---|---|
|  |  | 0 | 1 | 2 | 4 |
| Cable | Cable Swap | 93.9 | 96.1 | 96.5 | **96.6** |
| Capsule | poke | 88.6 | 90.8 | 92.0 | **92.2** |
| Pill | Combined | 87.9 | 92.6 | 93.3 | **94.4** |
| Bracket brown | Parts mismatch | 83.4 | 83.8 | 84.5 | **86.3** |
| Connector | Parts mismatch | 86.2 | 90.2 | 90.7 | **91.2** |
| Metal plate | Scratches | 98.8 | **99.9** | **99.9** | **99.9** |

*H. Anomaly Detection for Automotive Plastic Parts*

**Automotive Plastic Parts Dataset (APPD).** This section introduces the Automotive Plastic Parts dataset, collected in-house and designed to assess the performance of the PCSNet model in real-world applications. The dataset comprises 4082 normal training images, along with an additional 1749 normal images and 69 abnormal images for testing. All images within the dataset are standardized at a resolution of 256×256 pixels. Unlike existing datasets like MVTec, this dataset primarily focuses on detecting subtle anomalies in automotive plastic parts. Capturing such anomalies through conventional overall shots is challenging. To address this challenge, a multi-angle shooting approach was employed, capturing multiple images of different parts of each component (see Fig. 13 (a)). However, this approach introduces variations in lighting, perspective, and shape, making it more difficult to distinguish normal images from abnormal ones. Notably, the dataset exhibits significant inter-class differences between normal samples and minor-sized anomalies, posing a considerable challenge for anomaly detection. The images of anomalies are displayed in Fig. 13(b).

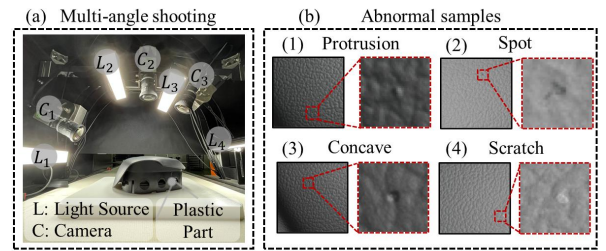

**Fig. 13.** (a) Schematic diagram of multi-angle photography; (b) Abnormal samples comprising four distinct anomaly categories.

**Anomaly detection results.** The results presented in Table VI demonstrate the superior performance of PCSNet compared to other methods. Specifically, in the 2-shot scenario, PCSNet shows a relative improvement of 16.6% over RD4AD and a remarkable enhancement of 29.8% and 27.8% over CFA and PatchCore, respectively. Moreover, with an increase in the number of training samples, PCSNet's performance continues to improve, achieving an impressive accuracy rate of 92.8% under the 10-shot condition. PCSNet outperforms the compared methods in anomaly detection due to its unique structural design and feature extraction capabilities. Unlike RD4AD and PatchCore, which rely solely on a pre-trained network for feature extraction, PCSNet employs a dedicated PFA sub-network that excels at capturing distinctive features specific to abnormal samples. This enables the model to overcome the challenge of capturing intricate details of small-sized anomalies. Similarly, CFA directly compares extracted features with normal prototypes as localization results without a fine-grain process, while PCSNet constructs a CAS sub-network for generating more accurate score maps with significantly fewer misclassifications. The localization results in Fig. 14 further demonstrate the superior accuracy of PCSNet in precisely identifying anomalies and reducing misclassifications of normal regions. These findings underscore the effectiveness of PCSNet in addressing the challenges posed by complex datasets containing subtle anomalies and significant variations among normal samples.

## V. CONCLUSION AND FUTURE WORK

In summary, this paper introduces a prototypical learning guided context-aware segmentation network (PCSNet), which effectively addresses the domain gap issue in few-shot anomaly detection. By incorporating a prototypical feature adaption (PFA) sub-network and a context-aware segmentation (CAS) sub-network, PCSNet offers a comprehensive solution to the challenge of capturing unique anomaly features. The PFA sub-network leverages contrastive learning and a pixel-level disparity classification loss function to bolster the discriminability of the sub-network. Meanwhile, the CAS sub-network comprehensively leverages contextual information to enhance anomaly detection. Experimental results substantiate the remarkable performance of PCSNet, surpassing existing FSAD methods and achieving state-of-the-art results across widely recognized MVTec and MPDD datasets. Notably, PCSNet also introduces a data augmentation strategy with a limited number of real abnormal samples, which significantly enhances anomaly detection performance. Furthermore, the practical applicability of PCSNet in real-world industrial settings is validated using an industrial automotive plastic parts dataset, tailored for the detection of subtle anomalies in automotive plastic parts. This underscores the practical applicability of PCSNet in real-world few-shot anomaly detection scenarios.

Nonetheless, the current information exchange between the PFA sub-network and CAS sub-network relies on a basic concatenation of similarity maps and extracted features. Future research avenues may delve into more sophisticated feature fusion techniques, such as attention mechanisms, to enhance the synergy between these two sub-networks. This holds the potential to further advance the capabilities of PCSNet in addressing complex and intricate anomaly detection challenges.

TABLE VI COMPARISON OF K-SHOT ANOMALY DETECTION RESULTS ON THE APPD DATASET: PRO (%).

| Method | 2 | 4 | 8 | 10 |
| --- | --- | --- | --- | --- |
| RD4AD | 67.4 | 68.4 | 77.1 | 74.7 |
| CFA | 54.2 | 54.9 | 62.3 | 84.3 |
| PatchCore | 56.2 | 62.9 | 61.6 | 75.5 |
| **PCSNet** | **84.0** | **89.8** | **88.6** | **92.8** |

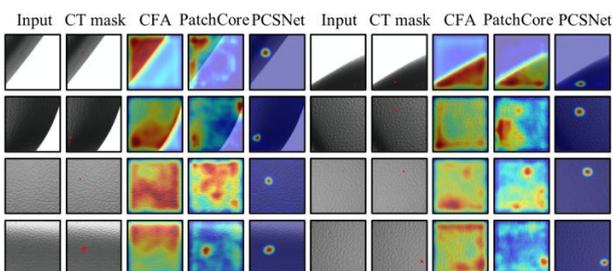

**Fig. 14.** The qualitative results of anomaly localization for PCSNet on APPD.